\documentclass[journal]{IEEEtran}

\ifCLASSINFOpdf

\else

\fi

\usepackage{amssymb}
\usepackage{graphicx}
\usepackage{subfigure}
\usepackage{float}
\usepackage{amsmath}
\usepackage{booktabs}
\usepackage{url}
\usepackage[numbers,sort&compress]{natbib}
\usepackage{array}
\usepackage{tabularx}

\begin{document}
\title{A Theoretical Study of The Relationship Between Whole An ELM Network and Its Subnetworks}

\author{Enmei Tu,
        Guanghao Zhang,
        Lily Rachmawati,
        Eshan Rajabally,
        Guang-Bin Huang
\thanks{Enmei Tu is with the Rolls-Royce@NTU Corporate Lab, Nanyang Technological University, Singapore}
\thanks{Guanghao Zhang and Guang-Bin Huang are with School of Electrical \& Electronic Engineering, Nanyang Technological University, Singapore}
\thanks{Lily Rachmawati is with Computational Engineering Team, Advanced Technology Centre,  Rolls-Royce Singapore Pte Ltd}
\thanks{Eshan Rajabally is with Future Technologies Group, Rolls-Royce Plc, UK}}
\markboth{}%
{Tu \MakeLowercase{\textit{et al.}}: A Relationship Between A  Network and Its Subnetworks}

\maketitle

\begin{abstract}
A biological neural network is constituted by numerous subnetworks and modules with different functionalities. For an artificial neural network, the relationship between a network and its subnetworks is also important and useful for both theoretical and algorithmic research, i.e. it can be exploited to develop incremental network training algorithm or parallel network training algorithm. In this paper we explore the relationship between an ELM neural network and its subnetworks. To the best of our knowledge, we are the first to prove a theorem that shows  an ELM neural network can be scattered into subnetworks and its optimal solution can be constructed recursively by the optimal solutions of these subnetworks. Based on the theorem we also present two algorithms to train a large ELM neural network efficiently: one is a parallel network training algorithm and the other is an incremental network training algorithm. The experimental results demonstrate the usefulness of the theorem and the validity of the developed algorithms.
\end{abstract}

\IEEEpeerreviewmaketitle

\section{Introduction}
Nowadays, huge volumes of data have been collected continuously in various fields, from engineering to scientific research. These data contain valuable information which usually appears in forms of complex patterns residing in the data and highly challenges most of current machine learning methods  (such as back propagation network \cite{li2012brief}) on  effectiveness and/or efficiency. Recently  manifold learning \cite{belkin2006manifold, tu2014novel, tu2016graph} and semisupervised learning \cite{hady2013semi, gong2015deformed, tu2013experimental, tu2015posterior} have drawn much attention due to their capability of learning some low dimensional distribution properties from high dimensional input space, but  researchers are still faced with imperative requirements of overcoming inefficiency or even incapability, due to insatiable memory and CPU demands, of training  models with huge amounts of data.

The Extreme Learning Machine (ELM) \cite{huang2006universal,huang2006extreme,huang2012extreme} was proposed as a single-hidden layer neural network for regression and classification problems due to its capability of universal approximation of almost any nonlinear or piecewise continuous function. The pivotal features of an ELM are that weights and bias of input-to-hidden layer (or input weight for short)  are randomly generated and no further tuning is required during the whole learning and prediction process. As a result, training a neural network is  reduced to training the hidden-to-output layer weight  (or output weight for short), which can be done by simply calculating the Moore-Penrose inverse of a hidden layer matrix.  Therefore, an ELM can achieve extremely fast training speed and meanwhile is able to attain a better generalization ability than other conventional methods \cite{huang2015extreme}. Moreover, extensive researches have shown the wide range of successful applications beyond just mathematical approximation, including human action recognition \cite{minhas2012incremental}, semi-supervised and unsupervised clustering \cite{huang2014semi}, image super resolution \cite{an2012image} and so on.


Although the ELM has shown strong capability for various research areas, scalability of big data learning is still a bottleneck of ELM method. Usually large amount of hidden neural nodes are required for complex pattern learning problems and consequently calculating Moore-Penrose inverse of large matrix directly is difficult and potentially impossible due to memory limitation. Training efficiency for a large volume of training data  may be another weakness even with a specific high performance toolbox such as \cite{akusok2015high}.

For a large scale ELM network learning problem, many researchers are devoted to developing ELM training methods to learn complex patterns from a large amount of data: Heeswijk et al \cite{van2011gpu} proposed a GPU-accelerated and parallelized method for big data learning. The main focus of this research is efficient learning with implementation on multiple GPU and CPU cores. An OS-ELM based ensemble classification method in super-peer P2P network \cite{sun2011elm} is proposed for online-sequential ELM training by similar intuition of parallelization training.  A high performance toolbox of ELM \cite{suri2015exploiting} focused on boosting training by CPU, GPU and HDF5 file format to achieve large scale training, fast file storage and easy installation. He et al \cite{he2013parallel} proposed a parallel ELM algorithm based on MapReduce, in which the matrix calculation for Moore-Penrose was decomposed and accelerated. A general framework based on MapReduce \cite{chen2016mr} is proposed by dividing the hidden layer into several groups, running a basic ELM training method for each group and then combining the output of all groups with same weight as the final output. An important concern of this method is that, in fact, simple combination output of each group is not theoretically equal to the basic ELM model with same number of hidden layer nodes . However, none of these researches study the relationship between an ELM network and its subnetworks. Here we show that this relationship is actually of great practical importance and can be exploited to develop algorithms for better training of an ELM network.

In this paper, we first prove a main theorem to reveal the relationship between an ELM network and its subnetworks. To the best of our knowledge, this is the first study to show that the optimal output weight of an ELM network is equal to a linear transformation of its subnetworks' optimal output weights. Based on this theorem, we also present two ELM training algorithms to train a large ELM network: a parallel network training algorithm and an incremental network training algorithm. We demonstrate the validity of the algorithms with experiments on four popular digits classification datasets.

The remainder of the paper is organized as follows: Section 2 briefly reviews key techniques of the ELM. Section 3 proves the main theorem of an ELM network and Section 4 describes two typical applications of the main theorem to solve a large ELM training problem. In Section 5, experimental results of the algorithms are presented, followed by discussions and conclusions in Section 6.

\section{A Brief Review of  Extreme Learning Machine}

The Extreme Learning Machine (ELM) method was proposed as a generalized multilayer feed-forward neural network with capability of classification and regression. In an ELM, the input weight and bias are randomly generated and then fixed through entire training and predicting process without any tuning.

Assume the training data set is $\{{{x}_{1}},{{x}_{2}},...,{{x}_{n}}\}$ and each sample is a $d$ dimensional vector. The corresponding target values of the training samples are $\{{{y}_{1}},{{y}_{2}},...,{{y}_{n}}\}$ and each target value is $c$ dimensional vector. For real value function regress $y_k$ is a real number.  For multiclass classification, $c$ is the number of classes and $y_k$ is a class indicator vector whose entries are all 0 except for that the $i^{th}$ entry is 1 if sample $x_k$ belongs to class $i$. Let us denote $X=\left( {{x}_{1}},{{x}_{2}},...,{{x}_{n}} \right)\in {{\mathbb{R}}^{d\times n}}$ and  $Y=\left( {{y}_{1}},{{y}_{2}},...,{{y}_{n}} \right)\in {{\mathbb{R}}^{n\times c}}$. For a multilayer feed-forward neural network with  $m$ hidden neurons, the network output corresponding to sample $x_i$ is
\begin{equation}\label{OneSampleF}
F({x_i})=\sum_{j=1}^m{w_j h_j({x_i})}={h}{(x_i)}{W}
\end{equation}
where ${W}=[w_1,w_2,....w_m]^T \in{{\mathbb{R}}^{m\times c}}$ is the output weight matrix\footnote{For simplicity, in the following parts we will mention $W$ as \emph{output weight} or \emph{optimal solution} of the network, depending on description context.}. ${h}({x_i})=[h_1({x_i}),h_2({x_i}),....h_m({x_i})]$ is a row vector representing hidden layer output of sample $x_i$, where $h(\cdot)$ is a continuous nonlinear function which maps samples from  $d$-dimensional input data space to $m$-dimensional feature space.
The mapping function $h(\cdot)$ is uniquely characterized by a random vector $a$, a random bias $b$ and an activation function $g$, i.e. for sample $x_i$ and hidden neuron $j$ the mapping function is  $h_j({x_i})=g({x_i;a_j,b_j})$.  Note that $F(x_i)$ is a row vector of length $c$. ELM theory has proven that if $g$ is a nonlinear continuous function \cite{huang2006universal} and $a$ and $b$ are randomly generated according to any continues probability distribution, then universal approximation property would be satisfied, which means that as the hidden layer neuron number  $m$ increases, the network can theoretically approximate \emph{any complex function} with sufficient accuracy.

Two most popular activation functions are sigmoid function and Gaussian function and respectively their expressions are
$$
g({x_i;a_j,}b_j)=\cfrac{1}{1+exp({x_i^Ta_j }+ b_j)}
$$
$$
g({x_i;a_j,}b_j)=exp(-b_j\lVert{x_i-a_j}\rVert)
$$
The input weight $a$ and bias $b$ are usually generated from uniform distribution [-1, 1].

Equation (\ref{OneSampleF}) is the output for one sample $x_i$ only. The outputs of  ELM with $m$ hidden neurons and $n$ input training samples are
\begin{equation}\label{ELMOutput}
\left\{ \begin{aligned}
& F({{x}_{1}})=\sum\limits_{j=1}^{m}{{{w}_{j}}{{h}_{j}}({{x}_{1}})}=h({{x}_{1}})W \\
& \quad \quad \ \ \vdots  \\
& F({{x}_{n}})=\sum\limits_{j=1}^{m}{{{w}_{j}}{{h}_{j}}({{x}_{n}})}=h({{x}_{n}})W \\
\end{aligned} \right.
\end{equation}
In matrix form, equation (\ref{ELMOutput}) can be written concisely as
$$
F=HW
$$
where
$$
{H}=\begin{bmatrix}
g({x_1;a_1,}b_1) & ... & g({x_1;a_m,}b_m) \\
\vdots&\vdots& \vdots \\
g({x_n;a_1,}b_1) & ... &
g({x_n;a_m,}b_m)
\end{bmatrix}
$$
is the hidden layer output matrix (or hidden layer matrix for short). $F$ is a matrix, in which row $i$ is $F(x_i)$.

Since the input weight and bias are randomly generated and fixed as constants, the output weight ${W}$ is the only parameter that needs to be tuned in network training process and can be obtained  by ridge regression with global optimality \cite{huang2012extreme}
\begin{equation}
\label{eq:optimization}
\underset{W\in {{\mathbb{R}}^{m\times c}}}{\mathop{\min }}\,{{\left\| F-Y \right\|}^{2}}+\alpha {{\left\| W \right\|}^{2}}
\end{equation}
where parameter $\alpha$ is the regularization parameter, representing the tradeoff between minimizing training error and   model generalization.

The analytic solution for the optimization (\ref{eq:optimization}) can be obtained by setting derivative of objective function  to zero, which yields
\begin{equation}
\label{eq:analyticsolution}
{W}=({H}^T{H}+\cfrac{{I}}{\alpha})^{-1}{H}^T{Y}
\end{equation}
where ${I}$ is the identity matrix.

For applications where there are more hidden neurons than training samples ($m>n$), though rare for big data learning, solutions for $W$ could be ill-conditioned. To handle this problem, Huang \emph{et al }\cite{huang2014semi} restrict $W$ to a linear combination of rows of ${H}$, i.e. ${W}={H^T\beta}$. In this case, ${HH^T}$ is invertible and by multiplying ${(HH^T)}^{-1}H$ on both side of the derivative of equation (\ref{eq:optimization}), the solution becomes
\begin{equation}\label{eq:analyticsolution2}
{{W}=H^T(HH^T+\cfrac{I}{\alpha})^{-1}Y  }
\end{equation}

\section{Relationship between ELM network and its subnetworks}
\begin{figure}[H]
	\centering
	\includegraphics[width=0.4\textwidth]{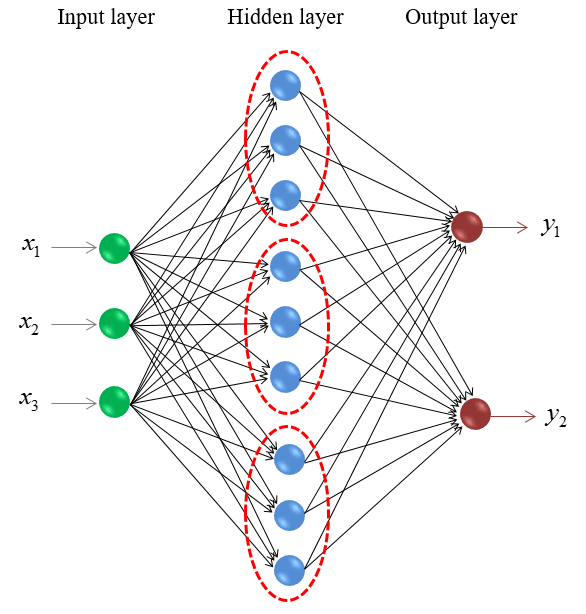}
	\caption{A neural network and its three subnetworks (indicated by red dash circles).}
	\label{SubNets}
\end{figure}

It has been demonstrated that biological neural networks contains numerous subnetworks, which have different functionalities and work in coordination to make the whole neural system functions optimally.  Therefore the relationships between the subnetworks and the whole neural system are of great importance in neural science and have become the most popular research topic in neural science.  For artificial neural networks, similar relationships between a network and its subnetworks are also important and useful for both theoretical and algorithmic research, because they can be used to study the properties of the network and to develop various training algorithms.  As an illustration, Fig. \ref{SubNets} displays a network and its subnetworks. However, as far as we know, this relationship has not been well studied. 
In this section we prove a theorem to show that an ELM network has a tight relationship with its subnetworks. In an ELM network, the input layer weight and bias are randomly generated.  Once they are fixed, the network structure is determined and its output layer weight is the only variable that needs to be learned during training process. Therefore, the relationship between a network and its subnetworks is primarily the relationship between their output weights.

For simplicity, let us consider a binary division case.  If the network is partitioned into two smaller networks, say network I and II, and their output weights are $W_1$ and $W_2$, respectively, our aim is to find out the relationship between the whole network output weight $W$ and the two smaller network output weights $W_1$ and $W_2$.

To be more specific, we will prove the following theorem for an ELM network:

\vspace{1em}
\emph{\textbf{Theorem 1}: If  $W_1$ and $W_2$ are the optimal output weights of two ELM networks and $W$ is the optimal output weight of an ELM network which is constructed by concatenating the two ELM networks together, then there exists a matrix $Z$ or  $\Delta W$ so that
	\begin{equation}\label{PELM1}
	W =Z\left[ \begin{matrix}
	{{W }_{1}}  \\
	{{W }_{2}}  \\
	\end{matrix} \right]
	\end{equation}
	or equivalently
	\begin{equation}\label{PELM2}
	W =\left[ \begin{matrix}
	{{W }_{1}}  \\
	{{W }_{2}}  \\
	\end{matrix} \right]-\Delta W
	\end{equation}}
Now we prove it and find out the exact analytic form of the matrix $Z$ and $\Delta W$.  Without loss of generality, let us assume that the hidden layer of a large network contains $2m$ neurons\footnote{In the rest parts of the paper, a network of size $m$ means  there are $m$ neurons in its hidden layer, since the neuron number in input layer and output layer is fixed for a given problem, i.e. they have to be equal to data dimension and classification classes (or function value dimension for regression), respectively.}, where $m$ is a positive integer. For manipulation simplicity, the network is partitioned into two equal subnetworks, and each subnetwork has $m$ hidden neurons\footnote{It should be mentioned that these assumptions are just to simplify expression. Our theorem and algorithms  are independent of the partitioning way of the whole network and the neuron number in each subnetwork. It neither requires  the network to have even number of neurons nor requires the network to be partitioned in equal size subnetworks.}. Accordingly, the hidden layer matrix $H$ can be written as a partitioned matrix $H=\left[ \begin{matrix}{{H}_{1}} & {{H}_{2}}  \\ \end{matrix} \right]$. Then for the case that number of training samples is greater than hidden layer neurons, i.e. $n>2m$,  we know from equation (\ref{eq:analyticsolution}) that the output weights of network I and II are
\begin{equation}\label{Beta1Beta2}
\left\{ \begin{aligned}
& {{W }_{1}}={{\left( \frac{{{I}_{m}}}{\alpha }+H_{1}^{T}{{H}_{1}} \right)}^{-1}}H_{1}^{T}Y \\
& {{W }_{2}}={{\left( \frac{{{I}_{m}}}{\alpha }+H_{2}^{T}{{H}_{2}} \right)}^{-1}}H_{2}^{T}Y \\
\end{aligned} \right.
\end{equation}
where $I_m$ is a $m\times m$ identity matrix.  Note that
$$
{{H}^{T}}H=\left[ \begin{matrix}
H_{1}^{T}  \\
H_{2}^{T}  \\
\end{matrix} \right]\left[ \begin{matrix}
{{H}_{1}} & {{H}_{2}}  \\
\end{matrix} \right]=\left[ \begin{matrix}
H_{1}^{T}{{H}_{1}} & H_{1}^{T}{{H}_{2}}  \\
H_{2}^{T}{{H}_{1}} & H_{2}^{T}{{H}_{2}}  \\
\end{matrix} \right]
$$
So the optimal output weight of the whole ELM network can be written as
\begin{equation}\label{EqBeta}
\begin{aligned}
& W={{\left( \frac{{{I}_{2m}}}{\alpha }+{{H}^{T}}H \right)}^{-1}}{{H}^{T}}Y \\
& \quad ={{\left( \frac{{{I}_{2m}}}{\alpha }+\left[ \begin{matrix}
		H_{1}^{T}{{H}_{1}} & H_{1}^{T}{{H}_{2}}  \\
		H_{2}^{T}{{H}_{1}} & H_{2}^{T}{{H}_{2}}  \\
		\end{matrix} \right] \right)}^{-1}}\left[ \begin{matrix}
H_{1}^{T}  \\
H_{2}^{T}  \\
\end{matrix} \right]Y \\
& \quad ={{\left( \left[ \begin{matrix}
		A & B  \\
		B^T & C  \\
		\end{matrix} \right] \right)}^{-1}}\left[ \begin{matrix}
H_{1}^{T}Y  \\
H_{2}^{T}Y  \\
\end{matrix} \right] \\
\end{aligned}
\end{equation}
where we let $A=\frac{{{I}_{m}}}{\alpha }+H_{1}^{T}{{H}_{1}}$, $B=H_{1}^{T}{{H}_{2}}$ and $C=\frac{{{I}_{m}}}{\alpha }+H_{2}^{T}{{H}_{2}}$. According to partitioned matrix inverse theory \cite{horn2012matrix}, a partitioned matrix inverse can be written as
$${{\left[ \begin{matrix}
		A & B  \\
		{{B}^{T}} & C  \\
		\end{matrix} \right]}^{-1}}=\left[ \begin{matrix}
{{S}_{C}}^{-1} & -{{S}_{C}}^{-1}B{{C}^{-1}}  \\
-{{C}^{-1}}{{B}^{T}}{{S}_{C}}^{-1} & D{{C}^{-1}}  \\
\end{matrix} \right]$$
if and only if submatrix $C$ and its Schur complement $S_C={\left( A-B{{C}^{-1}}{{B}^{T}} \right)}$ are both invertible, where $D=I+{{C}^{-1}}{{B}^{T}}{{S}_{C}}^{-1}B$. Noting that $\left( \frac{{{I}_{2m}}}{\alpha }+{{H}^{T}}H \right)$ is a positive definite matrix, these two conditions are naturally met according to the following lemma \cite{bernstein2009matrix}:

\vspace{1em}
\emph{\textbf{Lemma 1}: The following three statements are equivalent: (1) matrix $\left[ \begin{matrix}
	A & B  \\
	{{B}^{T}} & C  \\
	\end{matrix} \right]$ is positive definite; (2) $A$ and its Schur complement $S_A={\left( C-B^T{{A}^{-1}}{{B}} \right)}$  are both positive definite; (3) $C$ and its Schur complement $S_C={\left( A-B{{C}^{-1}}{{B}^{T}} \right)}$  are both positive definite.}
\vspace{1em}

So equation (\ref{EqBeta}) now becomes
\begin{equation}\label{EqBeta2}
\begin{aligned}
& W=\left[ \begin{matrix}
S_{C}^{-1} & -S_{C}^{-1}B{{C}^{-1}}  \\
-{{C}^{-1}}{{B}^{T}}S_{C}^{-1} & D{{C}^{-1}}  \\
\end{matrix} \right]\left[ \begin{matrix}
H_{1}^{T}Y  \\
H_{2}^{T}Y  \\
\end{matrix} \right] \\
& \quad =\left[ \begin{matrix}
S_{C}^{-1}H_{1}^{T}Y-S_{C}^{-1}B{{C}^{-1}}H_{2}^{T}Y  \\
-{{C}^{-1}}{{B}^{T}}S_{C}^{-1}H_{1}^{T}Y+D{{C}^{-1}}H_{2}^{T}Y  \\
\end{matrix} \right] \\
\end{aligned}
\end{equation}
Note that in equation (\ref{Beta1Beta2}), $W_1$ and $W_2$ are actually ${{A}^{-1}}H_{1}^{T}Y$ and ${{C}^{-1}}H_{2}^{T}Y$, respectively. Furthermore, since ${\left(AB\right)}^{-1}=B^{-1}A^{-1}$, we have ${{S_C}^{-1}}={{\left( A-B{{C}^{-1}}{{B}^{T}} \right)}^{-1}}={{\left( I-{{A}^{-1}}B{{C}^{-1}}{{B}^{T}} \right)}^{-1}}{{A}^{-1}}$. So equation (\ref{EqBeta2}) can be written as
\begin{equation}\label{Beta3}
W=\left[ \begin{matrix}
E{{W}_{1}}-S_{C}^{-1}B{{W}_{2}}  \\
-{{C}^{-1}}{{B}^{T}}E{{W}_{1}}+D{{W}_{2}}  \\
\end{matrix} \right]
\end{equation}
More obviously,
\begin{equation}\label{ParallelBeta1}
W =\left[ \begin{matrix}
E & -S_{C}^{-1}B  \\
-{{C}^{-1}}{{B}^{T}}E & D  \\
\end{matrix} \right]\left[ \begin{matrix}
{{W }_{1}}  \\
{{W }_{2}}  \\
\end{matrix} \right]
\end{equation}
where $E={{\left( I-{{A}^{-1}}B{{C}^{-1}}{{B}^{T}} \right)}^{-1}}={{S_C}^{-1}}A$. From equation (\ref{ParallelBeta1}) we can see that the matrix $Z$ in the theorem has the form
\begin{equation}
Z =\left[ \begin{matrix}
E & -S_{C}^{-1}B  \\
-{{C}^{-1}}{{B}^{T}}E & D  \\
\end{matrix} \right]
\end{equation}

Alternatively, from lemma 1 we know that the matrices $A$, $C$, $S_A$ and $S_C$ are all invertible, so the inverse of  the partitioned matrix can also be written as \cite{bernstein2009matrix}
\begin{equation}
{{\left[ \begin{matrix}
		A & B  \\
		{{B}^{T}} & C  \\
		\end{matrix} \right]}^{-1}}=\left[ \begin{matrix}
S_{C}^{-1} & -S_{C}^{-1}B{{C}^{-1}}  \\
-S_{A}^{-1}{{B}^{T}}{{A}^{-1}} & S_{A}^{-1}  \\
\end{matrix} \right]
\end{equation}
In this case equation (\ref{EqBeta}) now becomes
\begin{equation}
\begin{aligned}
& W=\left[ \begin{matrix}
S_{C}^{-1} & -S_{C}^{-1}B{{C}^{-1}}  \\
-S_{A}^{-1}{{B}^{T}}{{A}^{-1}} & S_{A}^{-1}  \\
\end{matrix} \right]\left[ \begin{matrix}
H_{1}^{T}Y  \\
H_{2}^{T}Y  \\
\end{matrix} \right] \\
& \quad =\left[ \begin{matrix}
S_{C}^{-1}H_{1}^{T}Y-S_{C}^{-1}B{{C}^{-1}}H_{2}^{T}Y  \\
-S_{A}^{-1}{{B}^{T}}{{A}^{-1}}H_{1}^{T}Y+S_{A}^{-1}H_{2}^{T}Y  \\
\end{matrix} \right] \\
\end{aligned}
\end{equation}
Similarly, substituting $W_1={{A}^{-1}}H_{1}^{T}Y$ and $W_2={{C}^{-1}}H_{2}^{T}Y$, we have
\begin{equation}\label{ParallelBeta2}
\begin{aligned}
& W =\left[ \begin{matrix}
S_{C}^{-1}A{{W }_{1}}-S_{C}^{-1}B{{W }_{2}}  \\
-S_{A}^{-1}{{B}^{T}}{{W }_{1}}+S_{A}^{-1}C{{W }_{2}}  \\
\end{matrix} \right] \\
& \quad =\left[ \begin{matrix}
S_{C}^{-1}A & -S_{C}^{-1}B  \\
-S_{A}^{-1}{{B}^{T}} & S_{A}^{-1}C  \\
\end{matrix} \right]\left[ \begin{matrix}
{{W }_{1}}  \\
{{W }_{2}}  \\
\end{matrix} \right] \\
\end{aligned}
\end{equation}
or
\begin{equation}
\label{MatZ}
Z =\left[ \begin{matrix}
S_{C}^{-1}A & -S_{C}^{-1}B  \\
-S_{A}^{-1}{{B}^{T}} & S_{A}^{-1}C  \\
\end{matrix} \right]
\end{equation}
From equation  (\ref{ParallelBeta1}) and (\ref{ParallelBeta2}) we can see the relationship between the whole network output weight and its subnetwork output weight: $W$ can be obtained by concatenating the subnetworks' output weight together and then multiplying a matrix to adjust it to be optimal. Here we actually also prove that a direct combination of subnetworks' output weight/output is not optimal, as in \cite{chen2016mr}, since matrix $Z$ is not equal to identity matrix.  Note that the adjustment matrix in equation (\ref{ParallelBeta1}) requires to compute  one matrix inverse $S_C^{-1}$ ($C^{-1}$ is already computed when solving $W_2$), but the adjustment matrix in equation  (\ref{ParallelBeta2}) requires to compute two matrices inverse (${S_A}^{-1}$ and ${S_C}^{-1}$) and this difference makes equation (\ref{ParallelBeta1}) more practical for design of efficient learning algorithms. Note that equation (\ref{MatZ}) can be further decomposed as
\begin{equation}
Z=\left[ \begin{matrix}
S_{C}^{-1} & O  \\
O & S_{A}^{-1}  \\
\end{matrix} \right]\left[ \begin{matrix}
A & -B  \\
-{{B}^{T}} & C  \\
\end{matrix} \right]
\end{equation}
where $O$ represents a zero matrix with proper size. From this equation we can see more obviously that instead of computing the inverse of the big matrix
$$\left[ \begin{matrix}
A & B  \\
{{B}^{T}} & C  \\
\end{matrix} \right]$$
in equation (\ref{EqBeta}),  we just need to compute two smaller matrix inverse $S_{C}^{-1}$ and $S_{A}^{-1}$ to obtain the optimal solution of the whole ELM network.

On the other hand, for the case $n<2m$, partition matrix $H$ in the same way but now we have
$$H{{H}^{T}}=\left[ \begin{matrix}
{{H}_{1}} & {{H}_{2}}  \\
\end{matrix} \right]\left[ \begin{matrix}
H_{1}^{T}  \\
H_{2}^{T}  \\
\end{matrix} \right]={{H}_{1}}H_{1}^{T}+{{H}_{2}}H_{2}^{T}$$
According to equation (\ref{eq:analyticsolution2}), the output weight now can be written as
\begin{equation}\label{Beta5}
\begin{aligned}
& W =\left[ \begin{matrix}
H_{1}^{T}  \\
H_{2}^{T}  \\
\end{matrix} \right]{{\left( \frac{I_{2m}}{\alpha }+{{H}_{1}}H_{1}^{T}+{{H}_{2}}H_{2}^{T} \right)}^{-1}}Y \\
& \quad =\left[ \begin{matrix}
H_{1}^{T}{{\left( A+{{H}_{2}}H_{2}^{T} \right)}^{-1}}Y  \\
H_{2}^{T}{{\left( C+{{H}_{1}}H_{1}^{T} \right)}^{-1}}Y  \\
\end{matrix} \right] \\
\end{aligned}
\end{equation}
where,  with some abuse of notation,  $A=\left( \frac{I_{2m}}{\alpha }+{{H}_{1}}H_{1}^{T} \right)$ and $B=\left( \frac{I_{2m}}{\alpha }+{{H}_{2}}H_{2}^{T} \right)$. Recall that the Woodbury inverse formula is
$$
{{\left( P+Q{{Q}^{T}} \right)}^{-1}}={{P}^{-1}}-{{P}^{-1}}Q{{\left( I+{{Q}^{T}}{{P}^{-1}}Q \right)}^{-1}}{{Q}^{T}}{{P}^{-1}}
$$
providing that $P^{-1}$ exists. Since both $A$ and $C$ are invertible according to \emph{lemma 1}, applying Woodbury formula to each submatrix in equation (\ref{Beta5})  we have
$$
W=\left[ \begin{matrix}
H_{1}^{T}\left( {{A}^{-1}}-{{A}^{-1}}{{H}_{2}}M_{1}^{-1}{{H}_{2}}^{T}{{A}^{-1}} \right)Y  \\
H_{2}^{T}\left( {{C}^{-1}}-{{C}^{-1}}{{H}_{1}}M_{2}^{-1}{{H}_{1}}^{T}{{C}^{-1}} \right)Y  \\
\end{matrix} \right]
$$
where for concise display purpose we denote ${{M}_{1}}=\left( I+{{H}_{2}}^{T}{{A}^{-1}}{{H}_{2}} \right)$ and ${{M}_{2}}=\left( I+{{H}_{1}}^{T}{{C}^{-1}}{{H}_{1}} \right)$. From equation (\ref{eq:analyticsolution2}) we know ${{W }_{1}}=H_{1}^{T}{{A}^{-1}}Y$ and ${{W }_{2}}=H_{2}^{T}{{A}^{-1}}Y$ are the output weights of subnetworks I and II, respectively. After substituting we have
$$
W=\left[ \begin{matrix}
{{W}_{1}}-{{H}_{1}}{{A}^{-1}}{{H}_{2}}M_{1}^{-1}{{H}_{2}}^{T}{{A}^{-1}}Y  \\
{{W}_{2}}-{{H}_{2}}{{C}^{-1}}{{H}_{1}}M_{2}^{-1}{{H}_{1}}^{T}{{C}^{-1}}Y  \\
\end{matrix} \right]
$$
Or more concisely,
\begin{equation}\label{Beta6}
W =\left[ \begin{matrix}
{{W }_{1}}  \\
{{W }_{2}}  \\
\end{matrix} \right]-\left[ \begin{matrix}
\Delta {{W }_{1}}  \\
\Delta {{W }_{2}}  \\
\end{matrix} \right]=\left[ \begin{matrix}
{{W }_{1}}  \\
{{W }_{2}}  \\
\end{matrix} \right]-\Delta W
\end{equation}
where $\Delta {{W}_{1}}={{H}_{1}}{{A}^{-1}}{{H}_{2}}M_{1}^{-1}{{H}_{2}}^{T}{{A}^{-1}}Y$ and $\Delta {{W}_{2}}={{H}_{2}}{{C}^{-1}}{{H}_{1}}M_{2}^{-1}{{H}_{1}}^{T}{{C}^{-1}}Y$. Equation (\ref{Beta6}) means that while $n<2m$, the whole network output weight can be obtained by concatenating its subnetworks' output weight and then subtracting an adjustment matrix. Note the the solution in equation (\ref{eq:analyticsolution}) is equivalent to that one in equation (\ref{eq:analyticsolution2}), so the two equations in \emph{Theorem 1} are also equivalent. \hfill$\square$

It should be mentioned that the final output weight in equation (\ref{ParallelBeta1}), (\ref{ParallelBeta2}) and (\ref{Beta6}) obtained by concatenating all output weights of its subnetworks is exactly same as the original ELM network. This guarantees the optimality of the solution, hence the performance of the network, as will be shown in the experiments in Section 5.

\section{Applications of The Theorem}
In this section we demonstrate the usefulness of the theorem. will develop two methods for training large scale ELM network using the relationship derived in previous section. The first method is a hierarchical algorithm and the second method is a block-wise incremental algorithm.

\subsection{Hierarchical Network Training Algorithm}
\begin{figure}[H]
	\centering
	\includegraphics[width=0.4\textwidth]{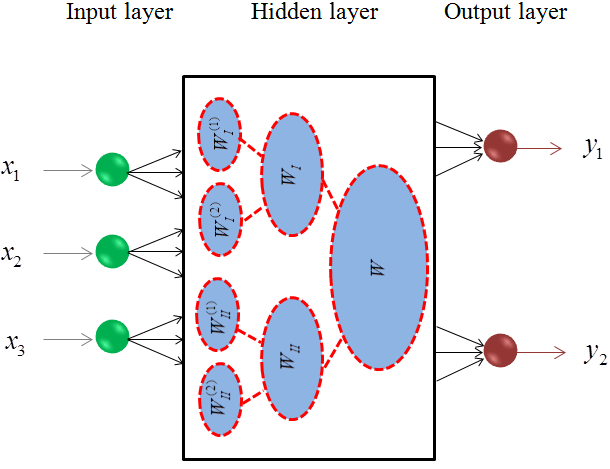}
	\caption{A two-level hierarchical training strategy (subnetworks are indicated by red dash circles).}
	\label{SubNets2}
\end{figure}

Equation (\ref{ParallelBeta2}) and  (\ref{Beta6}) tell us that in order to train a large network, we can first partition the network into smaller networks and train each subnetwork individually. Thereafter the output weights of all subnetworks can be used to construct the output weight of the whole network. This divide-and-train strategy can be easily implemented  in parallel to make use of hardware computational ability, i.e. on a multi-core computer or a cluster of computers. Meanwhile memory space requirement is also reduced,  because the training process only concerns  matrices with half size of the original problem  and the space requirements of matrix inverse is cubic in terms of matrix size. Furthermore, the divide-and-train strategy can be further applied to each subnetwork in a hierarchical way, i.e. to continue dividing each subnetwork into two further smaller networks and so on. As an illustration, Fig \ref{SubNets2} shows a two-level hierarchical division network.

From equation (\ref{ParallelBeta2}) we know that the relationship between the output weights are:
\begin{equation}\label{BetaLevel1}
W =Z\left[ \begin{matrix}
{{W }_{I}}  \\
{{W }_{II}}  \\
\end{matrix} \right]
\end{equation}
and
\begin{equation}\label{BetaLevel2}
\left\{ \begin{aligned}
& {{W }_{I}}={{Z}_{I}}\left[ \begin{matrix}
W _{I}^{(1)}  \\
W _{I}^{(2)}  \\
\end{matrix} \right] \\
& {{W }_{II}}={{Z}_{II}}\left[ \begin{matrix}
W _{II}^{(1)}  \\
W _{II}^{(2)}  \\
\end{matrix} \right] \\
\end{aligned} \right.
\end{equation}

Therefore, to obtain the optimal solution $W$ of the whole (potentially large) network, it is sufficient to train the four much smaller networks to get $W _{I}^{(1)}$, $W _{I}^{(2)}$, $W _{II}^{(1)}$ and $W _{II}^{(2)}$ and then use equations (\ref{BetaLevel2}) and (\ref{BetaLevel1}) to compute $W$ easily. Training a smaller network has at least three obvious advantages: (1) A smaller matrix manipulation is time and space saving; (2) A smaller matrix inverse tends to be more robust to disturbance and noise; (3) To make full use of  parallel architecture of multi-core computer or cluster of computers, the subnetworks can be implemented to run in parallel to further speedup training process.

\subsection{Incremental Network Training Algorithm}
\begin{figure}[H]
	\centering
	\includegraphics[width=0.4\textwidth]{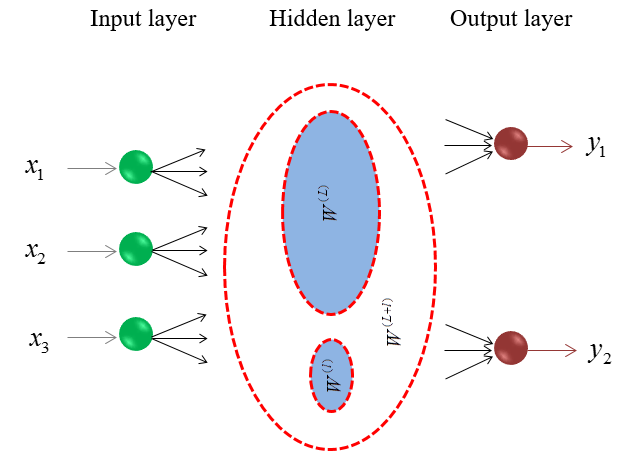}
	\caption{Incremental Network Learning (subnetworks are indicated by red dash circles).}
	\label{SubNets3}
\end{figure}

If subnetwork II in equation (\ref{PELM1}) or (\ref{PELM2}) is treated as a new added part, then the relationship can be utilized to train an incrementally growing network by just solving the new added subnetwork output weight and updating efficiently the whole network output weight. Suppose the network has $L$ neurons in hidden layer and it has already been trained to obtain its output weight $W^{(L)}$. Now $l$ neurons are added to the network and the whole network optimal output weight becomes $W^{(L+l)}$, as shown in Fig. \ref{SubNets3}. 

According to equation (\ref{PELM1}) we know that
\begin{equation}
W^{(L+l)} =Z\left[ \begin{matrix}
{{W }^{(L)}}  \\
{{W }^{(l)}}  \\
\end{matrix} \right]
\end{equation}
where $W^{(l)}$ is  obtained by using the same training data set to train the new added subnetwork of size $l$. More specifically, according to equation (\ref{ParallelBeta1}) we have
$$
W^{(L+l)} =\left[ \begin{matrix}
E{{W^{(L)} }}-S_{C}^{-1}B{{W^{(l)} }}  \\
-{{C}^{-1}}{{B}^{T}}E{{W^{(L)} }}+D{{W^{(l)} }}  \\
\end{matrix} \right]
$$
If we write
$$
P=\left[ \begin{matrix}
E  \\
-{{C}^{-1}}{{B}^{T}}E  \\
\end{matrix} \right],\quad
Q=\left[ \begin{matrix}
S_{C}^{-1}B  \\
-D  \\
\end{matrix} \right] $$
we have a more concise updating formula
\begin{equation}\label{UpdateELM}
{{W }^{(L+l)}}=P{{W }^{(L)}}-Q{{W }^{(l)}}
\end{equation}

The update equation (\ref{UpdateELM}) can be computed much more easily than using equation (\ref{eq:analyticsolution2}) to solve the whole network problem, because the new added neuron number $l$ is usually much smaller than the whole network hidden neuron number $L+l$. Matrices $C^{-1}$ and $S_C^{-1}$ can be calculated easily using previous solutions and thus both $P$ and $Q$ can be obtained efficiently\footnote{It should be mentioned that equation (\ref{UpdateELM}) can also be derived from equation (\ref{PELM2}), in which case that matrices $P$ and $Q$ are slightly different: $P=\left[ \begin{matrix} S_{C}^{-1}A  \\ -S_{A}^{-1}{{B}^{T}}  \\ \end{matrix} \right]$,  $Q=\left[ \begin{matrix} S_{C}^{-1}B  \\ -S_{A}^{-1}C  \\ \end{matrix} \right]$. But this update equation needs more computational cost, since at each updating time it requires to compute $S_A^{-1}$  which is of size $L$.}. A special case of equation (\ref{UpdateELM}) is that the new added subnetwork contains only one neuron in hidden layer ($l=1$). In this case the updating process can be implemented without explicitly computing any matrix inverse \cite{li2015inverse}.

\section{Experimental Results}
In this section we conduct experiments on four popular hand-writing digits datasets to demonstrate the validity of the theorem and the proposed hierarchical and incremental algorithms. The information of the experimental datasets are listed in Table \ref{DataInfo}.

\begin{table}[h]
	\centering  
	\footnotesize
	\caption{Inforamtion of the experimental datasets.}
	\label{DataInfo}
	\begin{tabular}{ c c c c c } \toprule \hline
		& usps      & mnist      &fontdigits      &pendigits        \\ \cmidrule(r){2-5}
		{sample \#}         & 9298      & 70000      &10000        &10992        \\ \cmidrule(r){2-5}
		{dimension \#}     & 256        & 784         &784            &16           \\ \cmidrule(r){2-5}
		{class \#}              & 10          & 10          &10             &10          \\ \hline \bottomrule
	\end{tabular}
	
\end{table}
The \emph{usps}\footnote{\url{https://www-i6.informatik.rwth-aachen.de/~keysers/Pubs/SPR2002/node10.html#tab:usps}} dataset contains  normalized $16\times 16$ grey scale image of US Postal Service handwritten digits. It has  7291 images for training and 2007 images for testing. The \emph{mnist}\footnote{\url{http://yann.lecun.com/exdb/mnist/}} dataset contains 70000 grey scale hand writing images, among them the first 60000 are training images and the rest 10000 are testing images. Each image is size-normalized and centered in a fixed-size image of size $28\times 28$. The \emph{fontdigits}\footnote{http://www.mathworks.com/help/nnet/examples/training-a-deep-neural-network-for-digit-classification.html} dataset contains 10000 grey scale digit images of size $28\times 28$, among them the first 5000 images are training images and the last 5,000 images are testing images. The \emph{pendigits}\footnote{\url{https://archive.ics.uci.edu/ml/datasets/Pen-Based+Recognition+of+Handwritten+Digits}} dataset consists of 10992 hand writing digits images. The first 7494 images are training images and the rest 3497 images are testing images. In each image $x$ and $y$ coordinates of each pixel on digit are normalized between $0...100$ to be the features.

In the hierarchical network training experiment, we only implement one-level hierarchical ELM network in figure \ref{SubNets2}, with each subnetwork contains 2000 hidden neurons.  As a comparison baseline algorithm, the original ELM network contains 4000 hidden neurons and is trained using equation (\ref{eq:analyticsolution}) or (\ref{eq:analyticsolution2}), depending on the relative relationship between the hidden neuron number and sample number. In incremental network training experiment, we start with an ELM with 2000 hidden neurons and then increase the hidden layer size by adding 2000 hidden neurons. The baseline algorithm, the original ELM, is also implemented to have an incremental hidden layer, but the output weight is computed  using equation (\ref{eq:analyticsolution}) or (\ref{eq:analyticsolution2}).   In order to test the robustness of the hierarchical and incremental network training methods, two most commonly used  activation functions are compared, i.e. the sigmoid function and the radial basis function. In each experiment, the algorithms run 5 times and the average of the results are used to compare their performance. The experimental results are in Table \ref{HExpResult} and Table \ref{IExpResult}, in which $W_H$ ($W_I$) is the optimal solution obtained by hierarchical (incremental) training and $W_O$ is the optimal solution obtained by original ELM. A similar notation is also utilized for the running time and error rate (i.e. TimeO and TimeH, ErrorO and ErrorH). AcFun stands for activation function.
\begin{table}[h]
	\centering  
	\footnotesize
	\caption{Hierarchical training experimental results}
	\label{HExpResult}
	\begin{tabular}{ c c c |c c} \toprule \hline
		& \multicolumn{2}{c}{fontdigits}      &\multicolumn{2}{c} {mnist}                   \\ \cmidrule(r){2-5}
		AcFun &                                                                        radbas&sigmoid &                                      radbas&sigmoid                                    \\ \cmidrule(r){1-5}
		{${{\left\| {{W}_{H}}-{{W}_{O}} \right\|}_{F}}$}         & 4.2e-27&4.3e-19                                  & 2.1e-27&3.3e-23                                      \\ \cmidrule(r){2-5}
		TimeO (sec)                                                                    & 3.46 &  3.37                                 &36.12 &  35.95                                           \\ \cmidrule(r){2-5}
		TimeH (sec)                                                                       & 2.48 &2.43                                  & 13.46&   12.87                                               \\ \cmidrule(r){2-5}
		ErrorO (\%)                                                                    &0.84  & 0.71                                 &6.79  & 3.40                                               \\ \cmidrule(r){2-5}
		ErrorH (\%)                                                                       & 0.84   &0.71                                  & 6.79  &3.40                                                \\ \hline
		\bottomrule
	\end{tabular}

	\bigskip
	\begin{tabular}{ c c c |c c  } \toprule \hline
		     &\multicolumn{2}{c}{usps}      &\multicolumn{2}{c}{pendigits}                \\ \cmidrule(r){2-5}
		AcFun &                                                                           radbas&sigmoid&                    radbas&sigmoid                     \\ \cmidrule(r){1-5}
		{${{\left\| {{W}_{H}}-{{W}_{O}} \right\|}_{F}}$}                                          &1.0e-26&4.4e-17                    &1.1e-29 &2.2e-15      \\ \cmidrule(r){2-5}
		TimeO (sec)                                                                                                 &4.88  &  4.91                  &8.83   & 4.95                 \\ \cmidrule(r){2-5}
		TimeH (sec)                                                                                               &2.90 &   2.94                  &5.62  &  3.06                   \\ \cmidrule(r){2-5}
		ErrorO (\%)                                                                                                 &5.23  &  4.53                       &2.05  &  2.57                 \\ \cmidrule(r){2-5}
		ErrorH (\%)                                                                                              &5.23  &  4.53                    &2.05   & 2.57                  \\ \hline
		\bottomrule
	\end{tabular}
\end{table}

\begin{table}[h]
	\centering  
	\footnotesize
	\caption{Incremental training experimental results}
	\label{IExpResult}
	\begin{tabular}{ c c c |c c  } \toprule \hline
		& \multicolumn{2}{c}{fontdigits}      &\multicolumn{2}{c} {mnist}                      \\ \cmidrule(r){2-5}
		AcFun &                                                                        radbas&sigmoid &                                      radbas&sigmoid                                      \\ \cmidrule(r){1-5}
		{${{\left\| {{W}_{H}}-{{W}_{O}} \right\|}_{F}}$}         & 4.0e-27&3.9e-19                                  & 2.0e-27&4.5e-23                                       \\ \cmidrule(r){2-5}
		TimeO (sec)                                                                    &3.62  &  3.44                                  & 35.32 &  37.92                                       \\ \cmidrule(r){2-5}
		TimeH (sec)                                                                       &2.65 &  2.662                                &12.86 &  13.50                                              \\ \cmidrule(r){2-5}
		ErrorO (\%)                                                                    &1.02  & 0.66                                 &6.81  &  3.43                                                \\ \cmidrule(r){2-5}
		ErrorH (\%)                                                                       &1.02  &  0.66                                  &6.81  &  3.43                                              \\ \hline
		
		\bottomrule
	\end{tabular}

	\bigskip

	\begin{tabular}{ c c c |c c  } \toprule \hline
		     &\multicolumn{2}{c}{usps}      &\multicolumn{2}{c}{pendigits}                \\ \cmidrule(r){2-5}
		AcFun &                                                                                              radbas&sigmoid&                    radbas&sigmoid                     \\ \cmidrule(r){1-5}
		{${{\left\| {{W}_{H}}-{{W}_{O}} \right\|}_{F}}$}                                      &1.0e-26&3.4e-17               &1.1e-29 &1.7e-15       \\ \cmidrule(r){2-5}
		TimeO (sec)                                                                                               &5.3575  &  5.0889                  &9.3865  &  5.69369        \\ \cmidrule(r){2-5}
		TimeH (sec)                                                                                                     &2.9692  &  3.1237                  &5.8048 &   3.6296               \\ \cmidrule(r){2-5}
		ErrorO (\%)                                                                                                 &5.48  &  4.58                      &2.07 &   2.49                 \\ \cmidrule(r){2-5}
		ErrorH (\%)                                                                                                     &5.48  & 4.58                    &2.07  &  2.49                  \\ \hline
		
		\bottomrule
	\end{tabular}
\end{table}

From these results we can see that for both two activation functions, the optimal solutions and performance of original ELM and hierarchical/incremental ELM are almost  identical\footnote{The third row is scientific number, i.e. 4.2e-27 is $4.2 \times 10^{-27}$.}. This demonstrates the correctness of the theorem and the validity of proposed training methods. It is worth mentioning that even if we only implemented one level hierarchical/incremental ELM here, the proposed network training methods is much efficient than original ELM, in terms of computational time and memory occupation, and show good potential to train a large ELM network. The larger the  network and dataset are, the more time and memory they can reduce, because the time and space complexity of the original ELM is cubic in terms of either network size in equation  (\ref{eq:analyticsolution}) or dataset size in equation  (\ref{eq:analyticsolution2}).


\section{Conclusions}
In this paper we study theoretically the relationship between an ELM network and its subnetworks. We prove a theorem which shows that the optimal solution of an ELM network is a linear transformation of its subnetworks' optimal solutions. This theorem has the potential  to be utilized to develop various efficient ELM training algorithms. As an example, we developed two algorithms to train a large ELM network: one is a hierarchical training algorithm and the other is an incremental training algorithm. The validity of both algorithms is demonstrated by experiments. For future work, we will focus on developing  more efficient algorithms for training large ELM networks based on the theorem and studying theoretically the criteria for recursively training subnetworks to construct a large ELM network.

\section*{Acknowledgment}
This work was conducted within the Rolls-Royce@NTU Corporate Lab with support from the National Research Foundation (NRF) Singapore under the Corp Lab@University Scheme.

\ifCLASSOPTIONcaptionsoff
  \newpage
\fi

\bibliographystyle{IEEEtran}
\bibliography{refs}

\end{document}